\RequirePackage{fix-cm}
\documentclass[smallextended]{svjour3}       
\smartqed  
\usepackage{graphicx}

\usepackage{times}
\usepackage{xcolor}
\usepackage{soul}
\usepackage[utf8]{inputenc}
\usepackage[small]{caption}
\usepackage{amsmath}
\usepackage{amsfonts}
\usepackage{amssymb}
\usepackage{booktabs}
\usepackage{pifont}
\usepackage{url}
\usepackage{natbib}
\usepackage{threeparttable}

\newcommand{\cmark}{\ding{51}}%
\newcommand{\xmark}{\ding{55}}%

\journalname{}
\usepackage{breakcites}

\begin{document}

\title{Dilated Temporal Relational Adversarial Network for Generic Video Summarization
}


\author{Yujia Zhang \and Michael Kampffmeyer \and Xiaodan Liang
        \and Dingwen Zhang \and Min Tan \and Eric P. Xing
}


\institute{Yujia Zhang \at
              Institute of Automation, Chinese Academy of Sciences; University of Chinese Academy of Sciences, 100190 Beijing, China. (Work done while the first author was at CMU)\\
              \email{zhangyujia2014@ia.ac.cn}           
           \and
           Michael Kampffmeyer \at
              Machine Learning Group, UiT The Arctic University of Norway, 9019 Troms{\o}, Norway.
           \and
           Xiaodan Liang \at
              Machine Learning Department, Carnegie Mellon University, 15213 Pittsburgh, USA.
           \and
           Dingwen Zhang \at
              Xidian University, 710071 Xi'an, China.
           \and
           Min Tan \at
              Institute of Automation, Chinese Academy of Sciences; University of Chinese Academy of Sciences, 100190 Beijing, China.
           \and
           Eric P. Xing \at
              Machine Learning Department, Carnegie Mellon University, 15213 Pittsburgh, USA.
}


\maketitle

\begin{abstract}
The large amount of videos popping up every day, make it more and more critical that key information within videos can be extracted and understood in a very short time. Video summarization, the task of finding the smallest subset of frames, which still conveys the whole story of a given video, is thus of great significance to improve efficiency of video understanding. We propose a novel Dilated Temporal Relational Generative Adversarial Network (DTR-GAN) to achieve frame-level video summarization. Given a video, it selects the set of key frames, which contain the most meaningful and compact information. Specifically, DTR-GAN learns a dilated temporal relational generator and a discriminator with three-player loss in an adversarial manner. A new dilated temporal relation (DTR) unit is introduced to enhance temporal representation capturing. The generator uses this unit to effectively exploit global multi-scale temporal context to select key frames and to complement the commonly used Bi-LSTM. To ensure that summaries capture enough key video representation from a global perspective rather than a trivial randomly shorten sequence, we present a discriminator that learns to enforce both the information completeness and compactness of summaries via a three-player loss. The loss includes the generated summary loss, the random summary loss, and the real summary (ground-truth) loss, which play important roles for better regularizing the learned model to obtain useful summaries. Comprehensive experiments on three public datasets show the effectiveness of the proposed approach.
\end{abstract}

\section{Introduction}
\label{intro}
Driven by the large number of videos that are being produced every day, video summarization~\citep{zhao2014quasi, sharghi2016query, meng2016keyframes} plays an important role in extracting and analyzing key contents within videos.
Video summarization techniques have recently gained increasing attention in an effort to facilitate large-scale video distilling~\citep{potapov2014category,zhang2016video,Mahasseni_2017_CVPR,plummer2017enhancing} due to its promising significance.
They aim to generate summaries by selecting a small set of key frames/shots in the video while still conveying the whole story, and thus can improve efficiency of key information extraction and understanding.

Essentially, video summarization techniques need to address two key challenges in order to provide effective summarization results: 1) how to exploit a good key-frame/key-shot selection policy that can take into account the long-range temporal correlations embedded in the whole video to determine the uniqueness and importance of each frame/shot; 2) from a global perspective, how to ensure that the resulting short summary can capture all key contents of the video with a minimal number of frames/shots, that is, how to ensure video information completeness and compactness.

Previous works have made some attempts toward solving these challenges. For instance, video summarization methods have to a large extent made use of Long Short-Term Memory (LSTM)~\citep{zhang2016video,Mahasseni_2017_CVPR}\citep{chen2017reference,chen2018show} and determinantal point process (DPP)~\citep{gong2014diverse,xu2015gaze,sharghi2016query} in order to address the first challenge and learn temporal dependencies. However, due to the fact that memories in LSTMs and DPPs are limited, we believe that there is still room to better exploit long-term temporal relations in the videos.

The second challenge is often addressed by utilizing feature-based approaches, i.e. instance motion features learning~\citep{zhao2014quasi,kim2014joint,gygli2015video}, to encourage diversity between the frames included in the summary. However, this cannot ensure the information completeness and compactness of summaries, leading to redundant frames and less informative results. 

Generative Adversarial Networks (GANs)~\citep{goodfellow2014generative} have been widely used in many computer vision tasks due to its effectiveness. Instead of only relying on the more traditional neural network approach that is only trained by Mean Squared Error (MSE) between the prediction and the ground-truth, the usage of GANs adds additional regularization. During training, the discriminator is encouraged to learn a complex loss function that encodes the higher order statistics of what a summary consists of, which in practice cannot be explicitly formulated by hand. A recent work~\citep{Mahasseni_2017_CVPR} utilizing adversarial neural networks reduces redundancy by minimizing the distance between training videos and the distribution of summaries, but it encodes all different information into one fixed-length representation, which reduces the model learning capabilities given different length of video sequences.

\begin{figure}[t]
\centerline{\includegraphics[width=8cm]{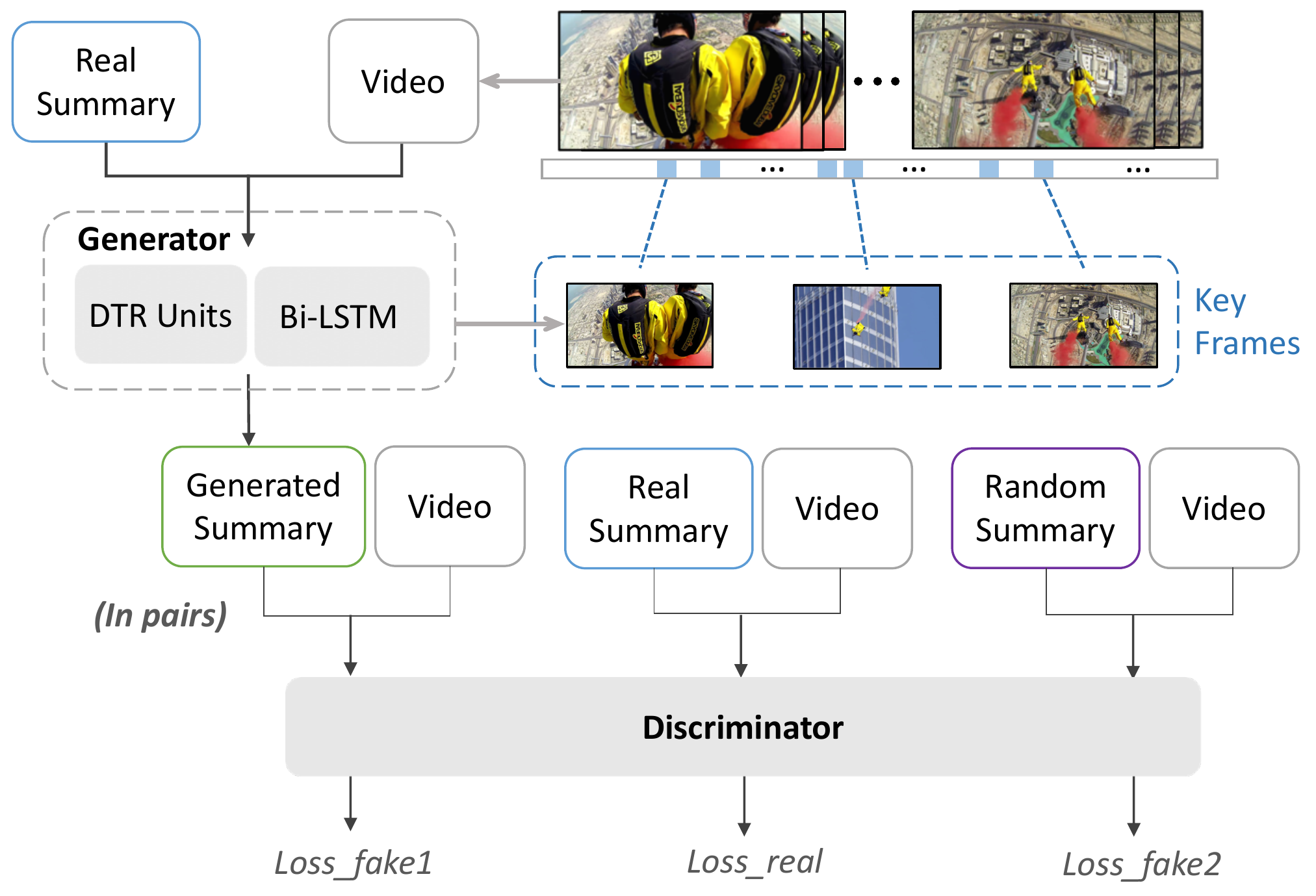}}
\caption{The proposed DTR-GAN aims to extract key frames which depict the original video in a complete and compact way. The DTR units are introduced to complement the commonly used Bi-LSTM, in order to better capture long-range temporal dependencies. The adversarial network with the supervised loss for the generator and the three-player discriminator loss, acts as a form of regularization to obtain better summarization results.}
\label{overview}
\end{figure}

To better address the above two core challenges in the video summarization task, namely modeling of long-range temporal dependencies and information completeness and compactness, we propose a novel dilated temporal relational generative adversarial network (DTR-GAN).
Figure~\ref{overview} shows an overview of the proposed method. The generator, which consists of Dilated Temporal Relational (DTR) units and a Bidirectional LSTM (Bi-LSTM)~\citep{graves2005framewise}, takes the real summary and the video representation as the input. DTR units aim to exploit long-range temporal dependencies complementing the commonly used LSTMs. The discriminator takes three pairs of input: generated summary pair, real summary pair and random summary pair and optimizes a three-player loss during training. To better ensure the completeness and compactness, we further introduce a supervised generator loss during adversarial training as a form of regularization.

Specifically, DTR units integrate context among frames at multi-scale time spans, in order to enlarge the model’s temporal field-of-view and thereby effectively model temporal relations among frames. We use three layers of DTR units, each modeling four different time spans, to capture short-term, mid-term and long-term dependencies. Bi-LSTM can function on every time step and benefit both long and short time dependencies by addressing the gradient problem commonly found in traditional non-gated Recurrent Neural Networks (RNNs)~\citep{graves2013speech}. Since our DTR units act on some certain time scales for efficiently capturing long and short temporal dependencies, Bi-LSTM can help with temporal modeling in parallel. In this way, combining DTR units with the LSTMs ensures that the generator can have better generating ability.

The discriminator takes three pairs of input: \emph{(generated summary, video sequences)}, \emph{(real summary, video sequences)} and \emph{(random summary, video sequences)}, and optimizes a three-player loss during training. It is cast to discriminate real summary from the generated summary, which further enhances the ability of the generator. At the same time, it ensures that the video representations are not learned from a trivial randomly shorten sequence. We further introduce a supervised generator loss during adversarial training to better ensure the completeness and compactness.

Our approach essentially achieves better model capability with DTR units by exploiting the global multi-scale temporal context. Further, the three-player loss-based adversarial network also provides more effective regularization to improve the discriminator's ability to recognize real summaries from fake ones. This, in turn, leads to better generated summaries. Evaluation results on three public benchmark datasets \emph{SumMe}~\citep{gygli2014creating}, \emph{TVSum}~\citep{song2015tvsum} and \emph{YouTube}~\citep{de2011vsumm} demonstrate the effectiveness of our proposed method.

In summary, this paper makes the following contributions:
\begin{itemize}
\item \textbf{DTR-GAN.} We propose a novel dilated temporal relational generative adversarial network for generic video summarization, which can generate a compact subset of frames with good information completeness and compactness. The experiments on three public datasets \emph{SumMe}, \emph{TVSum} and \emph{YouTube} demonstrate the effectiveness of the proposed approach.
\item \textbf{DTR units.} We develop a new temporal modeling module, Dilated Temporal Relational (DTR) unit to depict global multi-scale temporal context and complement the commonly used Bi-LSTM. DTR units dynamically capture different levels of temporal relations with respect to different hole sizes, which can enlarge the model's field-of-view to better capture the long-range temporal dependencies.
\item \textbf{Adversarial network with three-player loss.} We design a new adversarial network with a three-player loss for generic video summarization, which adds regularization to improve the model abilities during adversarial training. Different from the traditional two-player loss, we introduce a generated summary loss, a random summary loss and the real summary (ground-truth) loss, to better learn summaries as well as avoid trivial summary results.
\end{itemize}

A preliminary version of this method appeared in~\citep{zhang2019dtr}. Here we extend our work by: 1) placing our work into a broader context and providing a thorough literature background discussion, 2) providing a more thorough description of the methodology 3) extending the experimental evaluation to two additional datasets, namely the SumMe and YouTube datasets; and 4) including a thorough experimental analysis in form of ablation and visualization studies.

The rest of the paper is organized as follows. In Section 2, we review the related work. We present our proposed approach for video summarization in Section 3 and report and analyze the experimental results in Section 4. Finally, Section 5 draws conclusions and points to future research directions.

\section{Related Work}
\label{sec2}

\subsection{Video Summarization}
Recent video summarization works apply both deep learning frameworks and other traditional technique to achieve key frame/shot-level summarization, leading to a significant improvement on this task. For example, 
\citet{gygli2015video} formulated it as a subset selection problem and used submodular maximization to learn a linear combination of adapted submodular functions. In~\citep{xu2015gaze}, egocentric video summarization was achieved by using gaze tracking information (such as fixation and saccade). They also used submodular function maximization to ensure relevant and diverse summaries. \citet{zhao2014quasi} proposed onLIne VidEo highLIGHTing (LiveLight), which can generate a short video clip in an online manner via dictionary learning, thus it enables to start processing arbitrarily long videos without seeing the entire video. Besides, \citet{zhang2016context} also adopted dictionary learning using the methodology of sparse coding with generalized sparse group lasso to ensure retaining most informative features and relationships. They focused on individual local motion regions and their interactions between each other.

More recently, works using deep learning frameworks have been proposed and have achieved great progress. \citet{zhou2017reinforcevsumm} used a deep summarization network via reinforcement learning to achieve both supervised and unsupervised video summarization. They designed a novel reward function that jointly takes diversity and representativeness of generated summaries into account. \citet{ji2017video} formulated the video summarization as a sequence-to-sequence learning problem and introduced an attentive encoder-decoder network (AVS) to obtain key video shots. They used LSTMs for both encoder and decoder for exploring contextual information. \citet{zhang2016video} also used LSTM networks. They proposed a supervised learning technique by using LSTM to automatically select both keyframes and key subshots, which is complemented with DPPs for modeling inter-frame repulsiveness to encourage diversity of generated summaries. There are some other works on DPP. \citet{gong2014diverse} proposed sequential determinantal point process (seqDPP), which heeds the inherent sequential structures in video data and retains the power of modeling diverse subsets, so that good summaries possessing multiple properties can be created. In~\citep{zhang2016summary}, keyframe-based video summarization was performed by nonparametrically transferring structures from human-created summaries to unseen videos. They used DPP for extracting globally optimal subsets of frames to generate summaries. In~\citep{yao2016highlight}, a pairwise deep ranking model was employed to learn the relationship between highlight and non-highlight video segments, to discover highlights in videos. They designed the model with spatial and temporal streams, followed by the combination of the two components as the final highlight score for each segment.

Moreover, in~\citep{meng2016keyframes}, videos were summarized into key objects by selecting most representative object proposals which were generated from videos. Thus a fine-grained video summarization was achieved and what objects appear in each video can be told. Later, \citet{zatsushi2018viewpoint} built a summary depending on the users viewpoints, as a way of inferring what the desired viewpoint may be from multiple groups of videos. They took video-level semantic similarity into consideration to estimate the underlying users' viewpoints and thus generated summaries by jointly optimizing inner-summary, inner-group and between-group variances defined on feature representation.

More recently, the video summarization task was also performed by using vision-language joint embeddings. For example, \citet{Chu_2015_CVPR} exploited video visual co-occurrence across multiple videos by using a topic keyword for each video. They developed a Maximal Biclique Finding (MBF) algorithm to find shots that co-occur most frequently across videos.~\citet{plummer2017enhancing} trained image features paired with text annotations from both same and different domains, by projecting video features into a learned joint vision-language embedding space, to capture the story elements and enable users to guide summaries with free-form text input. \citet{panda2017collaborative} summarized collections of topic-related videos with topic keywords. They introduced a collaborative sparse optimization method with a half-quadratic minimization algorithm, which captures both important particularities arising in a given video and generalities arsing across the whole video collection.

\subsection{Generative Adversarial Networks}

Generative Adversarial Networks (GANs)~\citep{goodfellow2014generative} consist of two components, a generator network and a discriminator network with an adversarial learning. The generator works on fitting the true data distribution while confusing the discriminator, whose task is to discriminate true data from fake one.

Recently GANs have been used widely for many vision problems such as image-to-image translation~\citep{zhu2017unpaired}, image generation~\citep{reed2016generative,ghosh2016contextual}, representation learning~\citep{salimans2016improved,mathieu2016disentangling} and image understanding~\citep{radford2015unsupervised,liang2017dual}. For example, \citet{zhu2017unpaired} used cycle-consistent adversarial networks to translate images from source domain to target domain in the absence of paired examples. In~\citep{reed2016generative}, a text-conditional convolutional GAN was developed for generating images based on detailed visual descriptions, which can effectively bridge the characters and visual pixels.

To the best of our knowledge, the only existing GAN-based video summarization approach is~\citep{Mahasseni_2017_CVPR}. In their work, video summarization was formulated as selecting a sparse subset of video frames in an unsupervised way. In their work, they developed a deep summarizer network for learning to minimize the distance between training videos and the distribution of their summarizations. The model consisted of an autoencoder LSTM as the summarizer and another LSTM as the discriminator. Thus the summarizer LSTM was trained to confuse the discriminator, which forced the summarizer to obtain better summaries. 
It introduced GAN framework to address this task and has achieved good success. So inspired by this work, and also the good learning ability of GANs, we apply DTR-GAN method using a GAN-based architecture. The adversarial loss is used to formulate the regularization on the generator to get better summaries. Different from this work, we design a three-player loss that takes the random summary, generated summary and ground-truth summary into account, to provide better regularizations. Moreover, in our generator network, we also introduce DTR units which can enhance the temporal context representation.

\section{Our Approach}
\label{sec3}

The proposed DTR-GAN framework aims to resolve the key frame-level video summarization problem by jointly training in an adversarial manner. In the following sections, we first introduce the new dilated temporal relational (DTR) units. We then present the details of our DTR-GAN network with a novel three-player loss.

\subsection{Dilated Temporal Relation Units}
A desirable video summarization model should be capable of effectively exploiting the global temporal context embedded in future and past frames of the video in order to better determine the uniqueness and vital roles of each frame. We thus investigate how to achieve a good temporal context representation by introducing a new temporal relation layer. 

Prior works for temporal modeling often simply use various LSTM architectures to encode the temporal dynamic information in the video. However, models purely relying on the memory mechanism of LSTM units may fail to encode long-range temporal context, such as when video sequences exceed 1000 time steps. Moreover, redundant frames often appear in a small neighborhood of each frame. Besides modeling the long-term temporal changes in the video, as can be done using LSTM units, it is, therefore, important to further model local and multi-scale temporal relations to obtain compact video summaries.

\begin{figure*}[t]
\centerline{\includegraphics[width=0.96\textwidth]{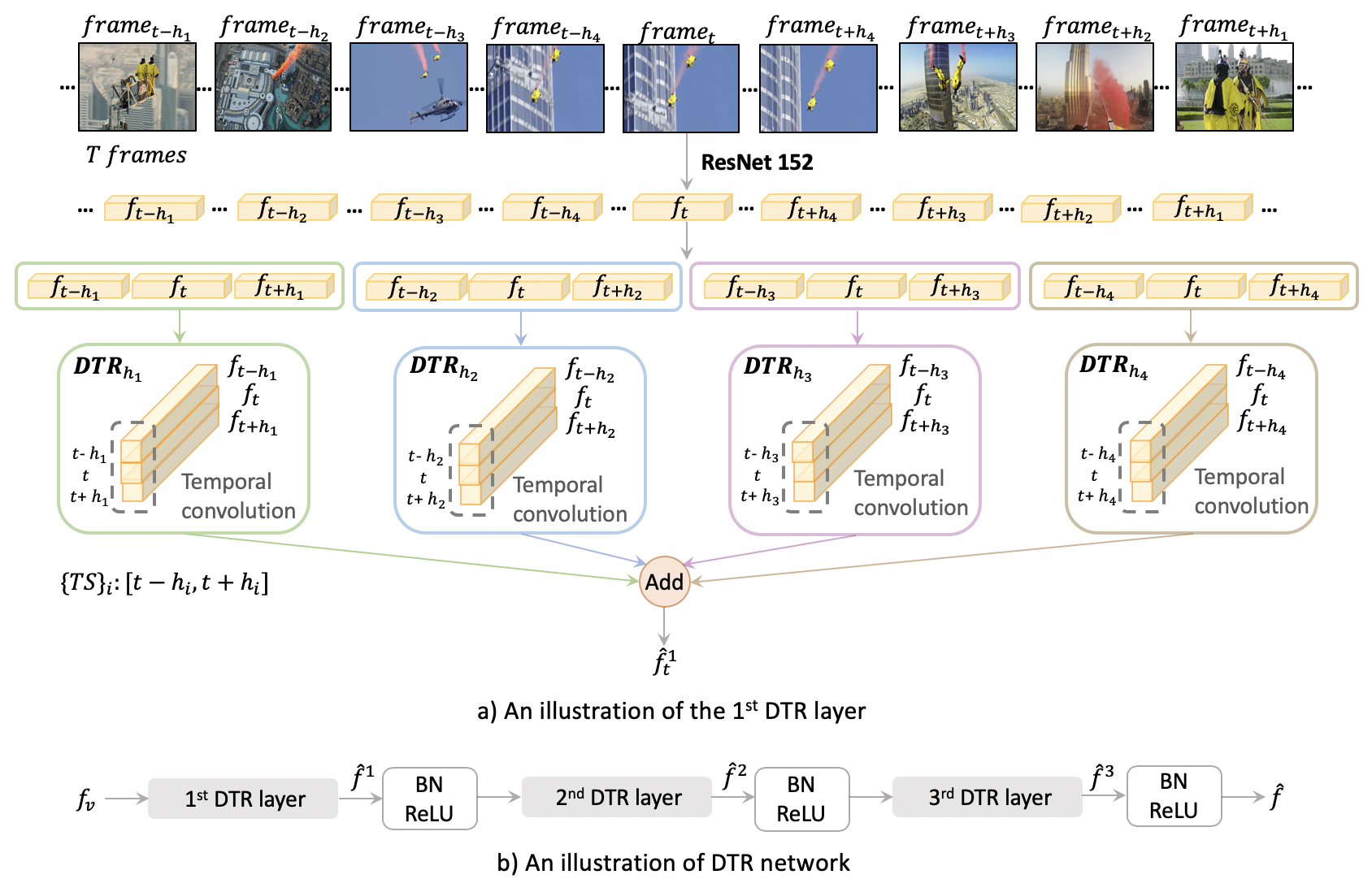}}
\caption{An illustration of the proposed Dilated Temporal Relational (DTR) unit. Given a video sequence with $T$ frames, where each frame has an appearance feature $f_t$, our DTR units dynamically capture different level of temporal relations by varying the hole sizes  $h_i$ for integrating temporal contexts from multi-range neighboring frames. As shown in (a), the $1^{st}$ DTR layer contains four DTR units with different hole sizes, each is a concatenation with temporal convolution. For a certain value of $h_i$, a new temporal relation range $\{TS\}_i$, ranging from $[t-h_i,t+h_i]$, is obtained. After that, a summation operation is used to merge the learned output together, as the output $\hat{f}_t^1$. The whole DTR network architecture is shown in (b). It takes the appearance features for all frames $f_v$ as the input and uses three DTR layers following a batch normalization layer and a relu layer for each DTR layer. After each DTR layer, learned representation $\hat{f}^1$, $\hat{f}^2$ and $\hat{f}^3$ are obtained. The final learned feature is illustrated as $\hat{f}$. By combining different temporal information from each DTR unit with respect to different $h_i$, we can enhance features of each frame by integrating multi-scale temporal contexts.}
\label{DTR}
\end{figure*}

Atrous convolutions have achieved great success for long-range dense feature extraction when employed in cascade or in parallel for multi-scale context capturing~\citep{chen2017rethinking} and for temporal convolution networks using a hierarchy of temporal convolutions~\citep{lea2016temporal}. Inspired by this, the key idea of our DTR unit is to capture temporal relational dependencies among video frames at multiple time scales. This is done by employing dilated convolutions across the temporal dimension, as illustrated in Figure~\ref{DTR}.

Given a certain video sequence ${V}=\{v_t\}_{t=1}^T$ of $T$ frames in total, we denote the appearance features of all frames as ${f_v}=\{f_t\}_{t=1}^T$. The features are extracted using the Resnet 152~\citep{he2016deep} model, which has been pretrained on ILSVRC 2015~\citep{russakovsky2015imagenet}.

Formally, DTR units function on the above appearance features $f_v$ of the whole video by incorporating temporal relations among frames in different time spans ${\{TS}\}_i$, corresponding to different hole sizes $h_i$. In our model, we use three DTR layers, each containing four different DTR units.

As shown in Figure~\ref{DTR}(a), for each frame $v_t$, a DTR layer enhances its feature $f_t$ using four DTR units, followed by the summation operation to merge all the information together, and generated the learned feature $\hat{f}_t^j$ in the $j^{th}$ layer. The enhanced feature $\hat{f}_t^j$ of the frame $v_t$ is computed as:

\begin{equation}
\hat{f}_t^j = \sum_{i=1}^M \mathbf{DTR}_{h_i}([f_{t-h_i}, f_t, f_{t+h_i}]),
\end{equation}

\noindent where $M$ denotes the number of different hole sizes used in each DTR layer, resulting in different $h_i$. Each $\mathbf{DTR}_{h_i}$ represents the transformation function that operates on the feature concatenation of $f_{t-h_i}$, $f_{t}$ and $f_{t+h_i}$, which has distinct parameters with respect to each hole size $h_i$. The transformation is formulated using a temporal convolution along temporal dimension only and results in a learned temporal representation feature of the same size as $f_t$. For each DTR unit, we empirically use $M=4$ and hole sizes $h_i$ of size 1, 4, 16, and 64. For each $h_i$, the time span for capturing temporal relations of each frame corresponds to:
\begin{equation}
    {\{TS\}}_i = 2\cdot h_i+1.
\end{equation}

In Figure~\ref{DTR}(b), an illustration of the DTR network with three layers of DTR units is shown. It takes the appearance feature of the video $f_v$ as the input. The output is defined as $\hat{f}^1$, $\hat{f}^2$ and $\hat{f}^3$ after different layers, following the batch normalization and ReLU operations, where $\hat{f}^j=\{\hat{f}_t^j\}_{t=1}^T$ at $j^{th}$ DTR layer. The final output of DTR network is defined as $\hat{f}$, where $\hat{f}=\{\hat{f}_t\}_{t=1}^T$, which combines different temporal relations among video sequences. After summing the features obtained from $\mathbf{DTR}_{h_i}$, the appearance feature $f_v$ can be converted into a temporal-sensitive feature $\hat{f}$ that explicitly encodes multi-scale temporal dependencies. The size of the filters are $\mathbf{\omega}\times W$, where $W$ corresponds to the filter size along the feature dimension. The size of receptive field can be computed as:

\begin{equation}
    RF = h_i\cdot(\mathbf{\omega}-1)\cdot j+1,
\end{equation}

\noindent where the size of the receptive field $RF$ is computed by different hole sizes $h_i$ at the $j^{th}$ layer. Here we use a filter $\mathbf{\omega}$ of size $3\times\/1$.

The DTR network expands the receptive field without any reduction in temporal resolution to model long-range temporal dependencies, which has the advantages over other spatio-temporal feature extractors, like~\citep{tran2015learning} and~\citep{feichtenhofer2016spatiotemporal}. In their work, they encode each video clip into a fixed descriptor and cannot produce the embedding results on a frame-level, which is required in our task for generating frame-level scores afterward.

Each DTR unit models the temporal relationships by capturing neighboring features of different time spans. Thus it can sense different neighboring features along the time space, and learns the dependencies among these different features. Besides, DTR also has the advantage of low computational complexity because of its simplicity. The proposed DTR unit is general enough to facilitate any network architectures to enhance temporal information encoding.

\subsection{DTR-GAN}
\subsubsection{Generator Network}
\begin{figure*}[t]
\centerline{\includegraphics[width=\textwidth]{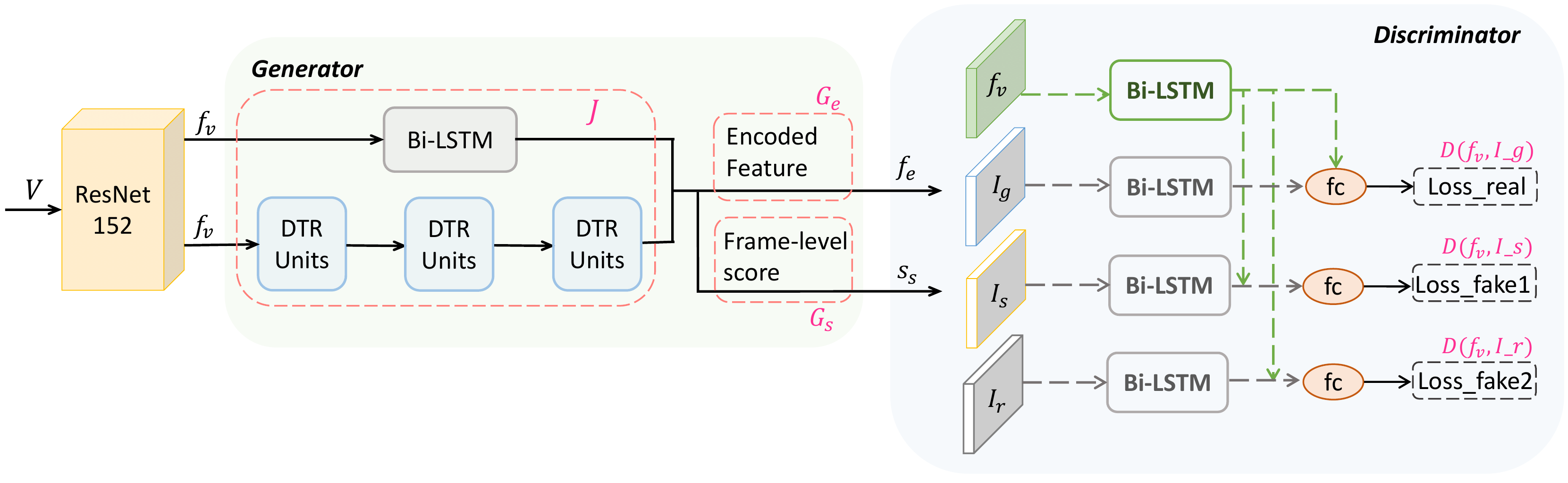}}
\caption{The network architecture of our DTR-GAN. Taking a video sequence $V$ as inputs, we can obtain appearance features of all frames $\mathbf{F}$ by passing the original frames into the pretrained ResNet-152 model. The generator $G$ is used to predict key frames consists of three components: 1) a Temporal Encoding Module $J$ that integrates both Bi-LSTM units and stacked DTR units is employed to produce enhanced features of each frame; 2) the confidence scores $s_s$ of all frames are then predicted by passing the features into the Summary predictor $G_s$; 3) as another branch, the enhanced features of all frames are combined into the features $f_e$ by the Compact Video Representation Module $G_e$. The discriminator $D$ then uses $f_e$ and $s_s$ to generate representations of three summaries, i.e. ground-truth summary $I_g$, predicted summary $I_s$ and randomly selected summary $I_r$. These three summary representations are then concatenated with the encoded features of the original video and are further fed into a shared Bi-LSTM module to get a real and two fake losses $D(f_v,I_g)$, $D(f_v,I_s)$, $D(f_v,I_r)$ in order to justify their fidelity.}
\label{framework}
\end{figure*}

As shown in Figure~\ref{framework}, given the appearance features $F$ of all frames, the generator $G$ aims to produce the confidence score $s_s$ of each frame being a key frame and the encoded compact video feature $f_e$. The whole generator architecture is composed of three modules: the temporal encoding module $J$ for learning the temporal relations among frames, the compact video representation module $G_e$ for generating the learned visual feature of the whole video, and the summary predictor $G_s$ for obtaining the final confidence score of each frame.

\textbf{a) Temporal Encoding Module $J$.} The module $J$ integrates a Bi-LSTM layer~\citep{graves2005framewise} and DTR network containing three DTR layers with twelve units in total, which encode both long-term temporal dependencies and multi-scale temporal relations with respect to different hole sizes.

The 2048-dimensional appearance features $\{f_t\}_{t=1}^T$ of all frames are taken as inputs of $J$. In the first branch, they are sequentially fed into one recurrent Bi-LSTM layer. The layer consists of both a backward and a forward path, each consisting of an LSTM with 1024 hidden cells, to ensure modeling of temporal dependencies both on past and future frames. We thus obtain an updated 2048-dimensional feature vector $\{\bar{f}_t\}_{t=1}^T$ for each frame as the concatenation of the forward and backward hidden states. 

In the second branch, following Eq.(1), each DTR layer computes $\hat{f}_t^j$ for each frame and passes it to the next DTR layer, and achieves multi-scale temporal relations among frames by making use of different hole sizes $h_i$ for better video representation. After passing over three DTR layers, we get the final evolved feature of each frame, and it is denoted as $\{\hat{f}_t\}_{t=1}^T$. Finally, the outputs of the module $J$ are two sets of updated features $\{\bar{f}_t\}_{t=1}^T$ and $\{\hat{f}_t\}_{t=1}^T$ for all frames.

\textbf{b) Compact Video Representation Module $G_e$.} Given the outputs of the module $J$, the encoded features of all frames are produced as $f_e = \{f^e_1, f^e_2, \dots, f^e_T\}$, where $f^e_t = G_e(\hat{f}_t,\bar{f}_t)$. In our setting, we use $G_e$ as a concatenation function followed by a fully connected layer, to learn the merged representation for video encoding.

The outputs of model $G_e$ denoted as $f_t^e$ are also used as the input of the discriminator network with three-player loss, which will be discussed later in Section~\ref{discriminator}.

\textbf{c) Summary Predictor $G_s$.} To predict confidence score $s_s = \{s_1, s_2, \dots, s_T\}$ for all frames as the video summary results, we introduce another summary predictor module $G_s$, as $s_t = G_s(\hat{f}_t,\bar{f}_t)$. The score is obtained by first concatenating of $\hat{f}_t$ and $\bar{f}_t$, and then passing the result to a fully-connected layer and a dropout layer that output one value for each input frame. After that, a sigmoid non-linearity is applied to each output value to produce the summary score. In this way, the confidence scores of all frames are generated by summary predictor $G_s$ as the final summary results.

\subsubsection{Discriminator Network}\label{discriminator}
In order to produce a high-quality summary, it is also desirable to evaluate whether the resulting summary encodes all main video contents of the original video and also consists of as few frames as possible from a global perspective. The key requirement is to measure the video correspondence between the obtained summary and the original video. 

Different from the traditional discriminator architecture~\citep{goodfellow2014generative} that only focuses on justifying the fidelity of a generated sample, the discriminator of our DTR-GAN instead learns the correspondence between input video and resulting summary, which can be treated as a paired target. Furthermore, in order to ensure that the summary is informative, we present a three-player loss. Instead of the commonly used two-player loss~\citep{arjovsky2017wasserstein,zhu2017unpaired}, this loss further enforces the discriminator to distinguish between the learned summary and a trivial summary consisting of randomly selected frames. The whole architecture is illustrated in Figure~\ref{framework}.

First, the inputs for the discriminator $D$ are three duplicates of the original video feature representation $f_v$, each paired with a different summary. The summaries are the ground-truth summary $I_g$, the resulting summary of the generator $I_s$, and a random summary $I_r$ respectively. The representation of each summary is obtained based on the feature representation $f_e$ from the generator, allowing the discriminator to utilize the encoded temporal information.

Let us denote the ground-truth summary score as $s_g\in \{0,1\}$, the resulting summary score as $s_s\in [0,1]$, and the random summary score, which is sampled from a uniform distribution, as $s_r\in [0,1]$. The dimension of $s_r$ is the same as the one of $s_g$ and $s_s$. The random summary score $s_r$ gives random importance scores for frames in the video. Then the summaries $I_g$, $I_s$ and $I_r$ can be computed by multiplying the corresponding encoded frame-level features ${f}_{e}$ with the summary scores $s_g$, $s_s$ and $s_r$, respectively:
\begin{align}
\begin{split}
&{I}_{g}={f}_{e}\cdot s_{g},\\
&{I}_{s}={f}_{e}\cdot s_{s},\\
&{I}_{r}={f}_{e}\cdot s_{r}.\\
\end{split}
\end{align}

The discriminator $D$ consists of four Bi-LSTM models, each with one layer, followed by a three-layer fully connected neural network and a sigmoid non-linearity to produce the discriminator score for the three pairs $(f_v,I_g)$,$(f_v,I_s)$ and $(f_v,I_r)$. All Bi-LSTMs have the same architecture but some of them have different parameters.

We pass the original video feature representation $f_v$ to one Bi-LSTM with a set of parameters, getting the hidden states, and pass the encoded summaries $I_g$, $I_s$ and $I_r$ to the other three Bi-LSTM with shared parameters, also getting the hidden states. The forward and backward paths in the Bi-LSTM consist of 256 hidden units each. We can thus obtain three learned representation pairs for checking the fidelity of the true representation pair and the other two fake ones. Then we concatenate each pair followed three fully connected layers. The dimensions of three layers are 512, 256 and 128. After that, a sigmoid layer is applied for obtaining the discriminator scores for each pair.

\subsubsection{Adversarial learning}
Inspired by the objective function proposed in recent work on Wasserstein GANs~\citep{arjovsky2017wasserstein}, which has good convergence properties and alleviates the issue of mode collapse, we optimize our adversarial objective with a three-player loss via a min-max game.

Specifically, given the three learned modules $J, G_e, G_s$ of the generator and the discriminator $D$, we jointly optimize all of them in an adversarial manner. The global objective over real loss $D(f_v,I_g)$ and the two fake losses $D(f_v,I_s)$, $D(f_v,I_r)$ ensures that the summaries capture enough key video representation, as well as avoids the learning of a trivial randomly shorten sequence as the summary. The min-max adversarial learning loss can be defined as:
\begin{align}
\begin{split}
     \underset{G}{\text{min}}&\underset{D}{\text{max}}\mathcal{L}(G,D) = {\mathbb{E}_g}[D({f}_{v},I_g)]\\& -\tau{\mathbb{E}_s}[D({f}_{v},I_s)] -(1-\tau){\mathbb{E}_r}[D({f}_{v},I_r)],
\end{split}
\end{align}
where $\tau$ is the balancing parameter between the resulting summary and the random summary. By substituting $J, G_e,G_s$ into $G$, and following Eq.(2), the objective can be reformulated as:
\begin{align}
\begin{split}
    & \underset{J, G_e,G_s}{\text{min}}\underset{D}{\text{max}}\mathcal{L}(J, G_e,G_s,D)\\&= {\mathbb{E}_g}[D({f}_{v},{G}_{e}(J(f_v))\cdot {s}_g)] \\
    & -\tau{\mathbb{E}_s}[D({f}_{v},{G}_{e}(J(f_v))\cdot {G}_{s}(J(f_v)))] \\
    & -(1-\tau){\mathbb{E}_r}[D({f}_{v},{G}_r(J(f_v))\cdot {s}_r)].
\end{split}
\end{align}

We empirically treat each player equally since both of the two fake pairs contribute to forcing the discriminator to learn the compact and complete real summary from fake ones. Thus we set the balancing parameter $\tau=0.5$, that is, $0.5$ for both of the fake pairs, namely the pairs of the generated summary and the random summary. 

To optimize the generator, we further incorporate a supervised frame-level summarization loss $\mathcal{L}_{summ}(G)$ between the resulting summary $s_s$ and the ground-truth summary $s_g$ during the adversarial training:
\begin{equation}
    \mathcal{L}_{summ}(G)=||s_s-s_g||^2_2.
\end{equation}

This loss aligns the generated summary with the real summary, guiding the generator to generate high-quality summaries by adding more regulations. The optimal generator can thus be computed as $G^*$:
\begin{equation}
    G^*=\text{arg}\underset{J, G_e,G_s}{\text{min}}\underset{D}{\text{max}}\mathcal{L}(J, G_e,G_s,D)+\mathcal{L}_{summ}(G).
\end{equation}

\subsection{Inference Process}

\begin{figure}[t]
\centerline{\includegraphics[width=8cm]{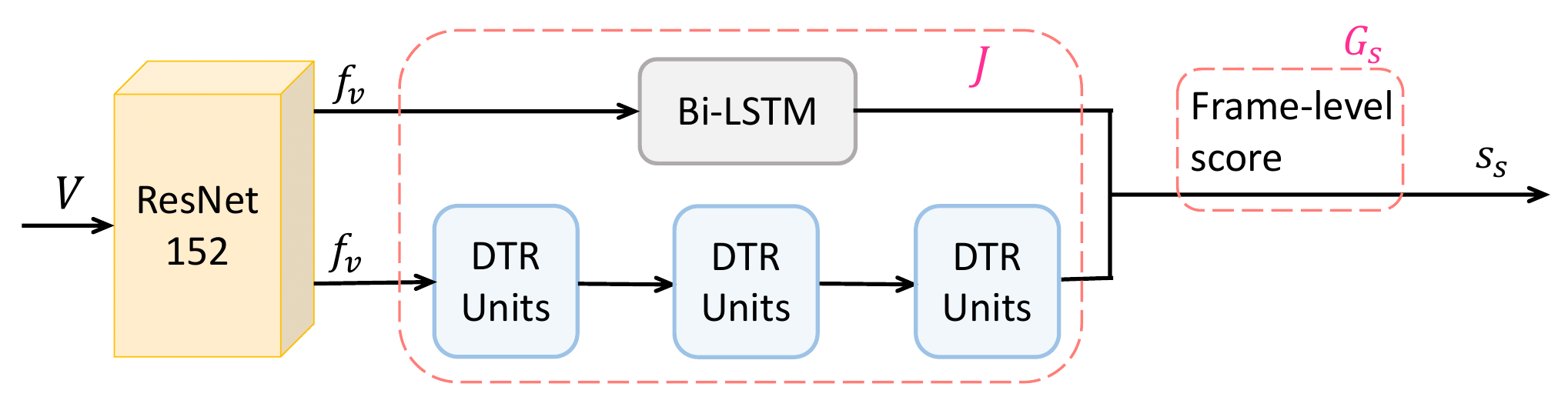}}
\caption{The inference process of the proposed DTR-GAN. The final confidence score for each frame of being key frame is obtained by passing the visual representation features to the temporal encoding module $J$ and the summary predictor $G_s$, without compact video representation model $G_e$ as well as the discriminator $D$.}
\label{inference}
\end{figure}

The inference process can be shown in Figure~\ref{inference}. Given each testing video, the proposed DTR-GAN model takes the whole video sequence as input. It then generates the confidence scores of all frames as the final summary result using only the generator during the inference process. Specifically, the testing video is first passed to the temporal encoding module $J$, generating the learned temporal representation, which can efficiently exploit global multi-scale temporal context. Then the summary predictor $G_s$ is applied to get the final predicted scores for each video.

Thus, the main differences for our DTR-GAN between training and inference phases are: 1) Discriminator $D$ is not used for inference, while training phase relies highly on it. 2) The compact Video Representation model, which is used to learn the merged video encoding for further training for discriminator $D$, is not required during inference phase.

\section{Experiments}
\label{sec4}
\subsection{Experimental Settings}
We will first introduce the three public datasets that we use and present the evaluation metrics for quantitative comparisons, before providing the details of the implementation.

\textbf{Datasets.} We evaluate our method on three public benchmark datasets for video summarization, i.e., \emph{SumMe}~\citep{gygli2014creating}, \emph{TVSum}~\citep{song2015tvsum} and \emph{YouTube}~\citep{de2011vsumm}. The SumMe dataset contains 25 videos covering multiple events from both the first-person and the third-person view. The length of the videos ranges from 1 to 6 minutes. The TVSum dataset contains 50 videos capturing 10 categories which are selected from the TRECVid Multimedia Event Detection (MED) task~\citep{smeaton2006evaluation}. It contains many topics such as news, cooking and sports and the length of each video ranges from 1 to 5 minutes. The YouTube dataset consists of 50 videos. The video lengths are from 1 to 10 minutes, and the contents include news, sports and cartoons. Following the previous methods~\citep{zhang2016video,Mahasseni_2017_CVPR}, we randomly select 80\% of the videos for training and 20\% for testing.

\textbf{Evaluation Metrics. } For fair comparison, we adopt the same keyshot-based protocol~\citep{zhang2016video} as in~\citep{Mahasseni_2017_CVPR} and \citep{zhou2017reinforcevsumm}, i.e., the harmonic F-measure, to evaluate our method, quantifying the similarity between the generated summary and the ground-truth summary for each video. Given the generated summary $A$ and the ground-truth summary $B$, the precision $P$ and recall $R$ of the temporal overlap are defined as:

\begin{align}
    \begin{split}
        & P=\dfrac{\text{overlap duration between }A \text{ and } B}{\text{duration of }A},\\
        & R=\dfrac{\text{overlap duration between }A \text{ and } B}{\text{duration of } B},
    \end{split}
\end{align}
the final harmonic F-measure $(F)$ is computed as:
\begin{equation}
    F=2P\cdot R/(P+R)\cdot 100\%.
\end{equation}

We also follow the process of~\citep{zhang2016video} to generate keyshot-level summaries from the key-frame level and the importance score-level summaries. We first apply the temporal segmentation method KTS~\citep{potapov2014category} to get video segments. Then if a segment contains more than one key frame, we give all frames within that segment scores of $1$. Afterward, we select the generated keyshots under the constraint that the summary duration should be less than 15\% of the duration of the original video by using the knapsack algorithm~\citep{song2015tvsum}.

\textbf{Implementation Details.}
We implement our work using the TensorFlow~\citep{abadi2016tensorflow} framework, with 1 GTX TITAN X 12GB GPU on a single server. We set the learning rate as 0.0001 for the generator and 0.001 for the discriminator. During the training process, we experimentally train the generator twice and train the discriminator once in each epoch. We randomly select a shot with 1000 frames and 10\% interval overlaps with neighboring shots to form each batch of the video in order to reduce the effect of edge artifacts. In test, we feed the whole video sequence as input, which can enable the model to sense the temporal dependencies in the whole time space.

\subsection{Comparison with the state-of-the-art methods}
We compare our DTR-GAN to the following supervised state-of-the-art methods to illus- trate the advantages of our algorithm:
\begin{itemize}
\item \pmb{Interestingness~\citep{gygli2014creating}} A method based on an interestingness score to select an optimal subset, which takes into account low-level information and high-level features.
\item \pmb{Submodularity~\citep{gygli2015video}} A method based on subset selection over summarization objectives: interestingness, representativess and uniformity (for retaining the temporal coherence).
\item \pmb{Summary transfer~\citep{zhang2016summary}} A non-parametric supervised approach that transfers the summary stuctures from human-created summaries of the training videos to unseen test videos.
\item \pmb{Seq-DPP~\citep{gong2014diverse}} A probabilistic model, sequential determinantal point process (seqDPP), for diverse sequential subset selection to select a subset of frames as summary result.
\item \pmb{DPP-LSTM~\citep{zhang2016video}} A method that exploits LSTMs to capture variable-range inter-dependencies, and uses DPP as an complement to encourage diverse selected frames.
\item \pmb{$\text{GAN}_\text{sup}$~\citep{Mahasseni_2017_CVPR}} A GAN-based method which aims to minimize the distance between feature representations of the training videos and their summarizations by integrating variational auto-encoders.
\item \pmb{$\text{DR-DSN}_\text{sup}$~\citep{zhou2017reinforcevsumm}} A deep summarization network based on deep reinforcement learning that jointly accounts for diversity and representativeness of generated summaries.
\item \pmb{$\text{DySeqDPP}$~\citep{li2018local}} Utilizes a dynamic Seq-DPP~\citep{gong2014diverse} together with a reinforcement learning algorithm to address the dynamic diverse subset selection problem, and to learn to impose the local diversity in the input videos.
\end{itemize}

Table~\ref{comparison of state-of-arts} shows the quantitative results on the ~\emph{SumMe},~\emph{TVSum} and \emph{YouTube} datasets.
It can be observed that our DTR-GAN substantially outperforms the other supervised state-of-the-art methods on three datasets. Particularly, on the SumMe dataset, DTR-GAN achieves 2.5\% better performance than the state-of-the-art method by~\citet{zhou2017reinforcevsumm} in terms of F-measure, and 3.2\% better on TVSum. Such performance improvements indicate the superiority of our DTR-GAN in encoding long-term temporal dependencies and correlations for determining the importance of each frame. At the same time, this also illustrates the effectiveness of validating the information completeness and summary compactness from a global perspective using our three-player adversarial training approach.

\begin{table}[t]
  \centering
  \renewcommand\arraystretch{1.2}
  \caption{Comparison results obtained by our method and other supervised approaches on \emph{SumMe}~\citep{gygli2014creating}, \emph{TVSum}~\citep{song2015tvsum} and \emph{YouTube~\citep{de2011vsumm}} datasets in terms of harmonic F-measure.}\label{comparison of state-of-arts}
  \begin{tabular}{c|c|c|c}
      \toprule
    \textbf{Method} & \textbf{SumMe} & \textbf{TVSum} & \textbf{YouTube}\\
        \midrule
        Interestingness~\citep{gygli2014creating} & 39.3 & - & -\\
    Submodularity~\citep{gygli2015video} & 39.7 & - & -\\
    Summary transfer~\citep{zhang2016summary} & 40.9 & - & 60.2\\
        Seq-DPP~\citep{gong2014diverse} & - & - & 60.8 \\
    DPP-LSTM~\citep{zhang2016video}  & 38.6 & 54.7 & -\\
    $\text{GAN}_\text{sup}$~\citep{Mahasseni_2017_CVPR} & 41.7 & 56.3 & 62.5\\
    $\text{DR-DSN}_\text{sup}$~\citep{zhou2017reinforcevsumm}  & 42.1 & 58.1 & {-}\\
    $\text{DySeqDPP}$~\citep{li2018local} & 44.3 & 58.4 & {-}\\
    \textbf{DTR-GAN} & \textbf{44.6} & \textbf{61.3} & \textbf{62.9}\\
    \bottomrule
  \end{tabular}
\end{table}

From Table~\ref{comparison of state-of-arts}, we can observe that our DTR-GAN achieves better performance (6.0\% and 6.6\% in terms of F-measure) than the DPP-LSTM work~\citep{zhang2016video} on two datasets. In \citep{zhang2016video}, the DPP-LSTM model is designed with containing two LSTM layers, one for modeling the forward direction video sequence, and the other for the backward direction. They also combine the LSTM layers’ hidden states and the input visual features with a multi-layer perceptron, together with the determinantal point process for enhancement. Thus, from the experimental results, we can come to the conclusion that DTR-GAN with LSTM and DTR networks can achieve better results by combining Bi-LSTM and DTR units together, allowing superior capturing of global multi-scale temporal relations.

Note that, another recent work~\citep{Mahasseni_2017_CVPR} also adopted the adversarial networks on temporal features produced by LSTMs for video summarization. However, our DTR-GAN is different from it: 1) the generator in~\citep{Mahasseni_2017_CVPR} encodes all different information into one fixed-length representation, which may reduce the model learning capabilities given different length of video sequence; 2) our DTR-GAN further introduces a new three-player loss to avoid that the network selects random trivial short sequences as the results; 3) in the generator network, besides the traditional LSTM, we further incorporate a new DTR unit to facilitate the temporal relation encoding by further exploiting multi-scale local dependencies.

The most recent state-of-the-art work by~\citet{zhou2017reinforcevsumm} achieves the best video summary result among the existing methods. The authors train deep summarization network based on LSTM networks via reinforcement learning. They design a reward function that jointly accounts for diversity and representativeness. In our work, we achieve 2.5\% and 3.2\% higher F-measure than~\citep{zhou2017reinforcevsumm}, due to the fact that it regularizes for generator in order to better obtain the summaries, as well as better temporal modeling by combining Bi-LSTM and DTR units.

\begin{table*}[t]
  \centering\renewcommand\arraystretch{1}
  \caption{Comparison of results for DTR units analysis for the \emph{SumMe~\citep{gygli2015video}} and \emph{TVSum~\citep{song2015tvsum}} dataset in terms of harmonic F-measure.}\label{dtr units}
  \renewcommand\arraystretch{1.2}
  \setlength{\tabcolsep}{2pt}
  \begin{threeparttable}
  \begin{tabular}{c|c|c|c}
      \toprule
    \textbf{Method} & \textbf{DTR units} & \textbf{SumMe} &\textbf{TVSum}\\
        \midrule
        \textbf{DTR-GAN} & \textbf{holes (1,4,16,64)} & \textbf{44.6} & \textbf{61.3}\\
    \hline
    DTR-GAN\_(holes 1,2,4,16) & holes (1,2,4,16) & 41.4 & 59.2\\
    \hline
    DTR-GAN\_(holes 16,32,64,128) & holes (16,32,64,128) & 42.6 & 60.8\\
    \bottomrule
  \end{tabular}
  \end{threeparttable}
\end{table*}

\subsection{DTR Units Analysis}
In this section, we analyze the summarization results when different hole sizes are used in the DTR units. The configurations that we are considering are:
\begin{itemize}
    \item \textbf{DTR-GAN\_(holes 1,2,4,16).} DTR units with hole size of (1,2,4,16) for each layer in order to compare the proposed hole size of (1,4,16,64) with this variant that uses a smaller range of temporal modeling.
    \item \textbf{DTR-GAN\_(holes 16,32,64,128).} DTR units with hole size of (16,32,64,128) for each layer in order to compare the proposed hole size of (1,4,16,64) with this variant that uses a larger range of temporal modeling.
\end{itemize}

As our DTR units employ different hole sizes to capture multi-scale temporal dependencies, it is also interesting to explore the effect of selecting different hole sizes on the summarization performance. We have tested two additional hole size settings, namely (1,2,4,16) and (16,32,64,128), whereas the proposed setting in all other experiments corresponds to (1,4,16,64). The model with hole size of (1,2,4,16) obtains a smaller range of temporal dilation, while the model with hole size of (16,32,64,128) achieves a larger range of temporal dilation, compared with the proposed DTR-GAN model, which contains intermediate a larger variants of hole sizes to capture multi-scale temporal relations better.

From Table~\ref{dtr units}, we can observe that both model variants achieve inferior performance to the 44.6\% and 61.3\% of DTR-GAN. Moreover, there is a minor performance difference between the results for larger and smaller dilation hole sizes.

The above comparison results indicate that with larger hole size we can obtain better results due to the larger time span. On the other hand, small holes are also required because of the fact that neighboring frames tend to share more similar features and have to some extent more temporal dependencies.

\begin{table*}[t]
  \centering\renewcommand\arraystretch{1}
  \caption{Comparison of results for our ablation experiments, indicating the importance of the various components in our model for the \emph{SumMe~\citep{gygli2015video}} and \emph{TVSum~\citep{song2015tvsum}} dataset in terms of harmonic F-measure. (The texts in blue color highlight the components that differ from the original DTR-GAN.)}\label{ablation analysis}
  \renewcommand\arraystretch{1.2}
  \setlength{\tabcolsep}{2pt}
  \begin{threeparttable}
  \begin{tabular}{c|c|c|c|c|c|c}
      \toprule
    \textbf{Method} & \textbf{DTR units} & \textbf{Bi-LSTM} & \textbf{G\_gt\_loss} & \textbf{Discriminator} &\textbf{SumMe} &\textbf{TVSum}\\
        \midrule
        \textbf{DTR-GAN} & \cmark \par & \cmark \par & \cmark \par & \textbf{3-player loss} & \textbf{44.6} & \textbf{61.3}\\
    \hline
    DTR-GAN w/o Bi-LSTM in G & \cmark \par & \color{blue}\xmark & \cmark \par & 3-player loss  & 43.7 & 59.6\\
    DTR-GAN w/o DTR units in G & \color{blue}\xmark & \cmark \par & \cmark \par & 3-player loss & 41.7 & 59.2\\
    \hline
    DTR-GAN w/o rand$^\text{1}$ & \cmark \par & \cmark \par & \cmark \par & \color{blue}2-player loss  & 40.6 & 59.4\\
    DTR-GAN\_LS loss$^\text{2}$ & \cmark \par & \cmark \par & \cmark \par & \color{blue}LS loss  & 42.9 & 60.2\\
    \hline
    DTR-GAN w G only & \cmark \par & \cmark \par & \cmark \par & \color{blue}\xmark  & 40.8 & 55.8\\
    \hline
    DTR-GAN w/o G\_gt\_loss & \cmark \par & \cmark \par & \color{blue}\xmark & 3-player loss & 41.9 & 56.9\\
    \bottomrule
  \end{tabular}
  \begin{tablenotes}
    \small
    \item[1] This one compares with the work~\citep{arjovsky2017wasserstein}
    \item[2] This one compares with the work~\citep{mao2017least}, and ``LS'' stands for ``least squares''
  \end{tablenotes}
  \end{threeparttable}
\end{table*}

\subsection{Ablation Analysis}
We conduct extensive ablation studies to validate the effectiveness of different components in our model by experimenting with different model variants. The different ablation analyses and the varied model component combinations on \emph{SumMe} and \emph{TVSum} datasets are as followed:

\subsubsection{Ablation Models}

\paragraph{Comparisons of Each Temporal Encoding Module}
\begin{itemize}
    \item \textbf{DTR-GAN w/o Bi-LSTM in G.} Drop the Bi-LSTM model in the generator in DTR-GAN to analyze the effect of the Bi-LSTM network in the proposed model.
    \item \textbf{DTR-GAN w/o DTR units in G}. Drop the DTR network in the generator in DTR-GAN to analyze the effect of the DTR units in the proposed model.
\end{itemize}

\paragraph{Comparisons of Disciminator}
\begin{itemize}
    \item \textbf{DTR-GAN w/o rand.} Apply two-player loss by dropping the random summary loss in the discriminator to analyze the effect of the three-player loss in the proposed model comparing with the commonly used two-player loss structure.
    \item \textbf{DTR-GAN\_least squares loss.} Apply the Least Squares loss function~\citep{mao2017least} instead of the loss designed in Wasserstein GAN~\citep{arjovsky2017wasserstein} to analyze the effect of loss functions in the proposed DTR-GAN.
\end{itemize}

\paragraph{Comparison of Adversarial Learning}
\begin{itemize}
    \item \textbf{DTR-GAN w G only.} Drop the discriminator part with the adversarial training, and use only the generator in order to analyze the effect of adversarial learning in the proposed model.
\end{itemize}
 
\paragraph{Comparison of Supervised Loss}
\begin{itemize}
    \item \textbf{DTR-GAN w/o G\_gt\_loss.} Drop the ground-truth loss in the generator to analyze the effect of the supervised loss for generating summaries with human annotated labels.
\end{itemize}

\subsubsection{Ablation Discussion}

In Table~\ref{ablation analysis}, we illustrate different settings including \emph{Bi-LSTM}, \emph{G\_gt\_loss} and \emph{Discriminator} components. As shown in the second row, the details of the proposed DTR-GAN are: hole sizes are 1,4,16, and 64 for DTR units in each DTR layer, Bi-LSTM and G\_gt\_loss are included, and three-player loss discriminator is applied. The rest rows show the different model variants for further ablation discussion, where the texts in blue color represent different components that differ from the proposed DTR-GAN model.

\textbf{The Effect of Each Temporal Encoding Module.} Note that in our generator network, we incorporate both the long-term LSTM units and multi-scale DTR units. By comparing model variants without either Bi-LSTM unit or our DTR unit with our full DTR-GAN, we can better demonstrate the effect of each module on the final summarization performance. It can be observed that the module capability is decreased by either removing Bi-LSTM units or DTR units.

From Table~\ref{ablation analysis}, we can see that by removing Bi-LSTM module, the performance of our approach decreases by 0.9\% and 1.7\%. While by removing DTR units the performance decreases by 2.9\% and 2.1\%. This shows that our DTR units have more effect than the Bi-LSTM module and this is due to the fact that better multi-scale temporal dependency helps learn better video temporal representation resulting in more compact and complete summaries. Besides, it also shows that Bi-LSTM can enhance the performance for the whole model, so we combine these two models together for better video summarization generation.

\textbf{The Effect of the Discriminator.}
We also test the performance of a model variant that only uses the standard two-player loss, i.e. the pairs of the original video with the ground-truth summary and with the generated summary. This is to validate the effectiveness of our proposed three-player objective in the adversarial training, which is also based on the Wasserstein GAN structure~\citep{arjovsky2017wasserstein}. We can observe that there is a large performance difference between standard two-player loss and our proposed three-player loss. The reason is that the random summary provides more regularization which ensures that the video representations are not learned from a trivial randomly shorten sequence.

Moreover, we replace the Wasserstein GAN with the Least Squares GAN~\citep{mao2017least} structure with our proposed three-player loss. From Table~\ref{ablation analysis}, we can see that the performance of this baseline is 42.9\% and 60.2\%, which are still 0.8\% and 2.1\% better than the results of previous state-of-the-art work by~\citet{zhou2017reinforcevsumm}. This further demonstrates that our proposed approach does not rely on GAN structure.

\textbf{The Effect of the Adversarial Learning Module.}
In addition, we also trained the model only using the generator. The performance of this baseline is only 40.8\% and 55.8\%, which are lower than most other ablation models and are 3.8\% and 5.5\% lower than the proposed DTR-GAN architecture. This demonstrates that the adversarial training with discriminator works better than non-adversarial training.

The discriminator functions to discriminate the ground-truth summary from generated and random summaries, which helps to enforce that the generator generates more complete and compact summaries. 

\textbf{The Effect of the Supervised Loss.}
During the adversarial training, we introduce the ground-truth loss for the generator as a form of regularization, by aligning the generated frame-level importance scores with the ground-truth scores.

From Table~\ref{ablation analysis}, we can see that this model obtains better performance on frame-level video summarization with the supervised loss. Specifically, by removing the \emph{``G\_gt\_loss''} component, the performance drops by 2.7\% and 4.4\%. This illustrates that our model can learn much better by using the human annotated labels.

\subsection{Qualitative Results}
\begin{figure*}[t]
\centerline{\includegraphics[width=\textwidth]{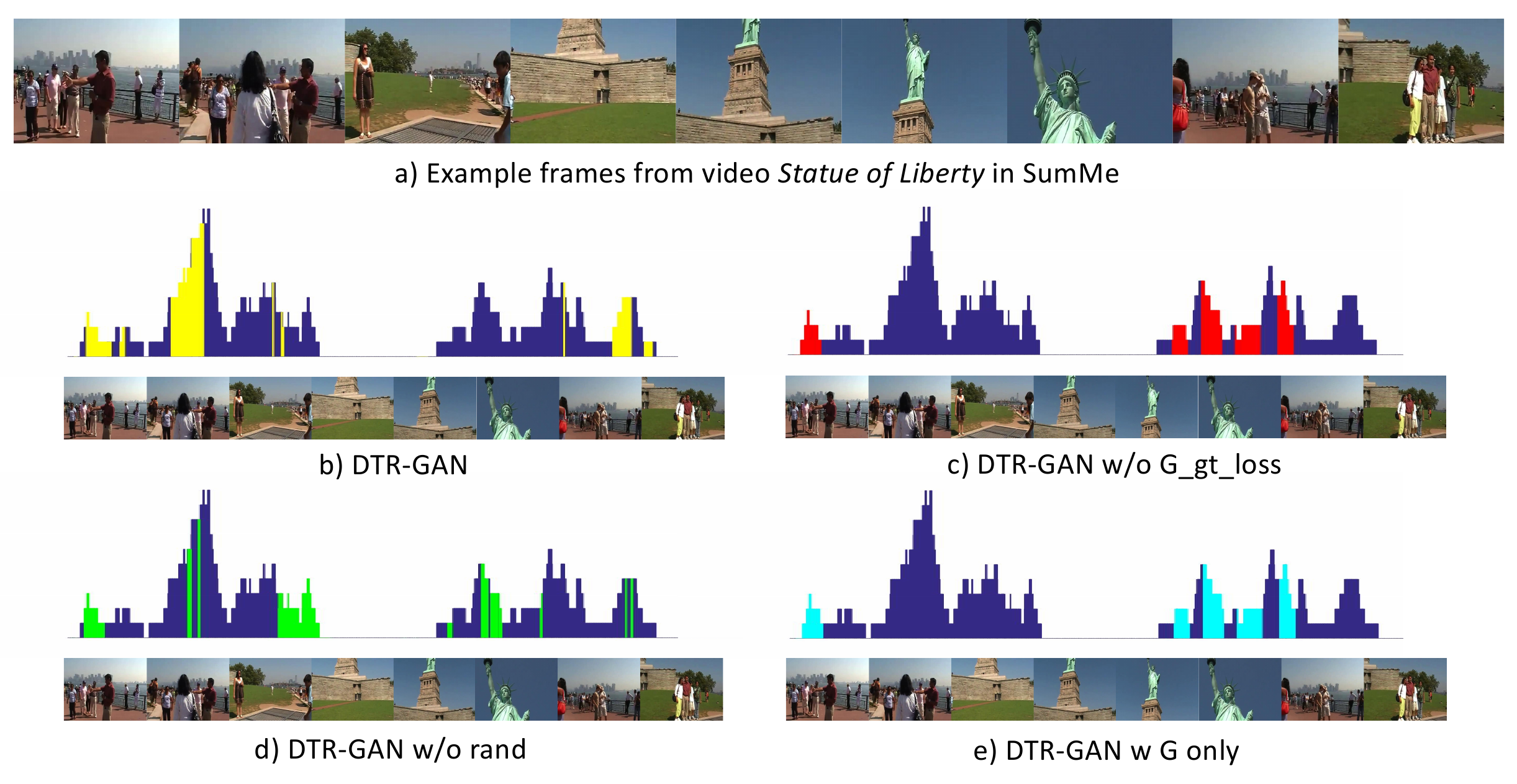}}
\caption{Video summarization results of some variants of our proposed DTR-GAN method for the video \emph{Statue of Liberty} in \emph{SumMe}~\citep{gygli2015video}. The dark blue bars in b), c), d), e) are the ground-truth frame-level scores, and the colored segments are the summary results generated by different model variants.}
\label{fig5}
\end{figure*}

\begin{figure*}[t]
\centerline{\includegraphics[width=\textwidth]{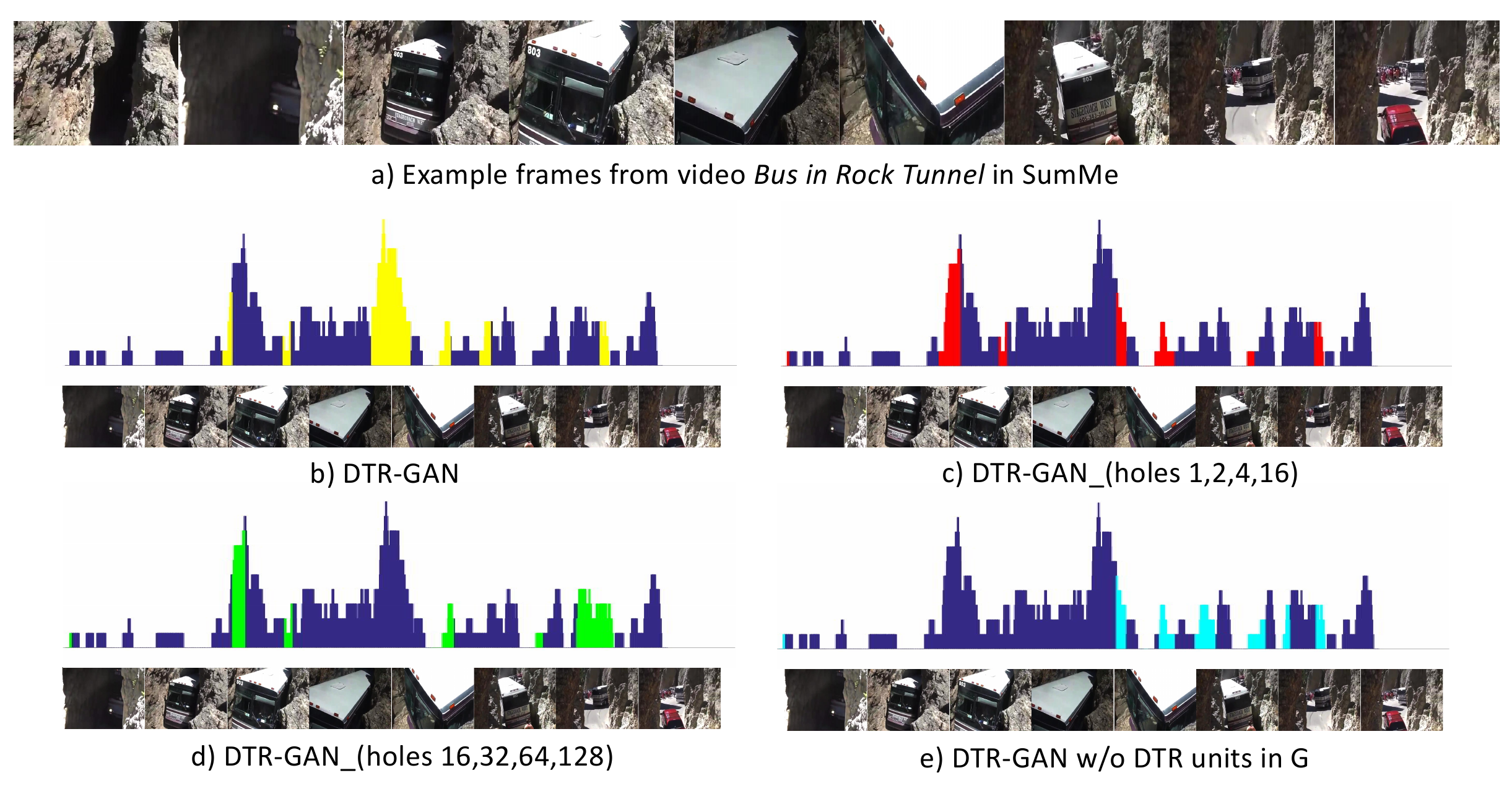}}
\caption{Video summarization results of some variants of our proposed DTR-GAN method for the video \emph{Bus in Rock Tunnel} in \emph{SumMe}~\citep{gygli2015video}. The dark blue bars in b), c), d), e) are the ground-truth frame-level scores, and the colored segments are the summary results generated by different model variants.}
\label{fig6}
\end{figure*}

To better demonstrate some key components of our framework, we visualize an example of the summary results overlaying the ground-truth frame-level important scores in Figure~\ref{fig5} and Figure~\ref{fig6}. 
We use the selected key frames obtained via the importance scores that are generated by the generator as a summary.

Figure~\ref{fig5} illustrates the visualized results on the video \emph{Statue of Liberty} in the \emph{SumMe} dataset on \emph{``DTR-GAN''}, \emph{``DTR-GAN w/o range''}, \emph{``DTR-GAN w/o G\_gt\_loss''} and \emph{``DTR-GAN w G only''}. Figure~\ref{fig6} illustrates the visualized results on the video \emph{Bus in Rock Tunnel} in the \emph{SumMe} dataset on \emph{``DTR-GAN''}, \emph{``DTR-GAN\_(holes 1,2,4,16)''}, \emph{``DTR-GAN\_(holes 16,32,64,128)''}, \emph{``DTR-GAN w/o DTR units in G''}.

From these figures, we can see that visualized results comply with the quantitative results in Table~\ref{ablation analysis}, where our model obtains reasonably better generated video summary results than the rest three models. All of the key components of our proposed framework contribute to improving overall performance.

\begin{figure*}
\centerline{\includegraphics[width=\textwidth]{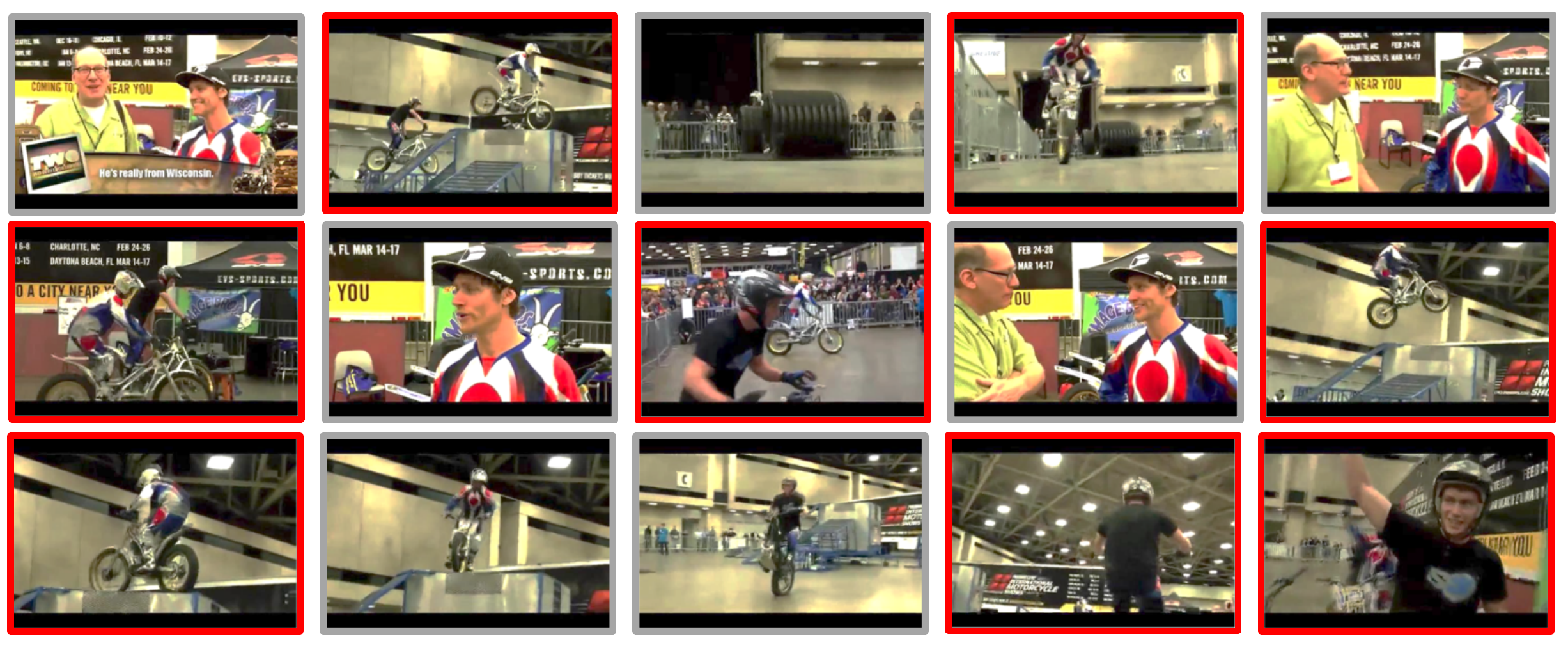}}
\caption{Qualitative examples of our proposed DTR-GAN for frame-level video summarization in \emph{TVSum} dataset. The images of red outlines are the selected key frames, while the images with grey outlines denote frames that were not selected.}
\label{exp_}
\end{figure*}

To further demonstrate the effectiveness of our framework, we provide qualitative examples of the summary results as shown in Figure~\ref{exp_}. The selected key frames are the images outlined in red, which are obtained via the importance scores that are generated by the generator as a summary. The images with grey outlines denote frames that were not selected.

From this figure, we can observe that the model tends to exclude the trivial information (i.e. interview parts in the video), and select more informative frames of the bicycle stunt show. At the same time, the bicycle stunt part is also summarized in order to better present the essence of the show, by removing redundant frames.

\section{Conclusion}
\label{sec5}
In this paper, we proposed DTR-GAN for frame-level video summarization. It consists of a DTR generator and a discriminator with three-player loss and is trained in an adversarial manner. Specifically, the generator combines two temporal dependency learning modules, Bi-LSTM and our proposed DTR network with three layers of four different hole sizes in each layer for multi-scale global temporal learning. In the discriminator, we use a three-player loss, which contains the generated summary, random summary, and ground-truth to introduce more restrictions during adversarial training. This helps the generator to generate more complete and compact summaries. Experiments on three public datasets \emph{SumMe}, \emph{TVSum} and \emph{YouTube} demonstrate the effectiveness of our proposed framework. In future work, we will continue to investigate this line of research by utilizing reinforcement learning algorithm~\citep{fu2019attentive}, attention mechanism~\citep{ji2019video} and multi-stage learning~\citep{huang2019user} within the DTR-GAN framework to further improve generic video summarization.

\begin{acknowledgements}
We would like to thank Xiaohui Zeng for her valuable discussions. This project is supported by the Department of Defense under Contract No. FA8702-15-D-0002 with Carnegie Mellon University for the operation of the Software Engineering Institute, a federally funded research and development center. This work is also partially funded by the National Natural Science Foundation of China (Grant No. 61673378 and 61333016), and Norwegian Research Council FRIPRO grant no.\ 239844 on developing the \emph{Next Generation Learning Machines}.
\end{acknowledgements}

\bibliographystyle{spbasic}      
\bibliography{main}   

\end{document}